\documentclass[times, 10pt,twocolumn]{article} 
\usepackage{latex8}
\usepackage{times}
\usepackage{graphicx}
\usepackage{subfigure}

\begin{document}

\title{Biologically Inspired Hierarchical Model for  Feature Extraction and Localization}

\author{Liang Wu\\ Institute for Brain and Neural Systems\\ Brown Univeristy, 182 Hope St., Providence, RI, 02912, USA\\ Liang\_Wu@brown.edu\\
}

\maketitle

\begin{abstract}
Feature extraction and matching are among central problems of computer vision. It is inefficent to search features over all locations and scales. Neurophysiological evidence shows that to locate objects in a digital image the human visual system employs visual attention to a specific object while ignoring others. The brain also has a mechanism to search from coarse to fine. In this paper, we present a feature extractor and an associated hierarchical searching model to simulate such processes. With the hierarchical representation of the object, coarse scanning is done through the matching of the larger scale and precise localization is conducted through the matching of the smaller scale. Experimental results justify the proposed model in its effectiveness and efficiency to localize features. 
\end{abstract}

\Section{Introduction}
In computer vision, there are two central problems: the extraction of robust features and the subsequent precise localization of those features. It has an expensive computational cost to search features at every location and scale. It is common to use interest operators \cite{operator}, salient feature detectors \cite{Kadir}, or selective visual attention \cite{attention} to select features for learning. However, matching is still computationally expensive given that the number of key points is usually over 10,000. \cite{SIFT}.

In this paper, we use a different approach placing the emphasis on efficient matching in a top-down attention manner. The model is inspired by human perception. When we search for an object in an image, keeping in mind roughly what the object looks like, we jump from one region to another with attention to some aspects of that object while ignoring others. From these aspects, we can finally localize the object. 
To build such a model in computer vision, a hierarchical representation of an object is given at many scales. Coarse scanning is done through the matching of the larger scale and precise localization is conducted through the matching of the smaller scale.

\section{Feature Description}
For feature extraction, we use Gabor filters for the reason that Gabor filters have been shown to be good simulations of visual cortex and they give a sparse representation of images \cite{Lee,gabor1}. 

2D Gabor kernels are characterized by the following equation,
\begin{eqnarray}
\psi_{f_0,\theta,\sigma}(x,y)=\frac{1}{\sqrt{2\pi}\sigma}e^{-\frac{4(x\cos\theta+y\sin\theta)^2+(y\cos\theta-x\sin\theta)^2)}{8\sigma^2}} \nonumber \\
\sin(2\pi f_0(x\cos\theta+y\sin\theta)).
\end{eqnarray}
where there are 3 parameters. Spatial frequency $f_0$ and scale $\sigma$ can be combined using frequency bandwidth $\phi$ with $2\pi f_0 \sigma =2\sqrt{ln2}(2^\phi+1)(2^\phi-1)$ \cite{Lee}. Since the frequency bandwidths of simple and complex cells have been found to range from $0.5$ to $2.5$ octaves centered around 1.2 and 1.5 octaves\cite{octave1,octave2}. We set $\phi=1.5$ in our system. 

\begin{figure}
\centering{
\includegraphics[width=.4\textwidth]{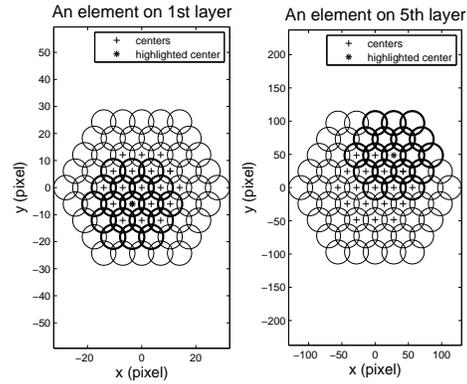}
\caption{Two elements on 2 layers.}}
\label{fig_sim}
\end{figure}

Note that in Gabor kernel we use $\sin$ only with $0$ phase shift. Different phases are crudely approximated by convolving the image with the kernel at all locations. Maximum operation is applied to make the detection tolerant to image distortions. The pooling range is only within each receptive field whose size is $2\sqrt{2}\sigma$. 
It is much smaller than the range used in other articles\cite{poggio} so that we do not lose much information about the locations of features which can be recovered in the subsequent matching. We do not take max operation across scale so that we do not lose scale information either.

In our system, receptive fields (RFs) are arranged similarly on $M$ different layers. The ratio of the size of each RF on a larger layer to that of its nearest smaller layer is kept at $\sqrt{2}$. We use this arrangement to capture features at different scales which later will also serve as an efficient way of matching.

There are $19$ feature detectors on each layer which are called elements. Each element is composed of $19$ adjacent RFs. As an illustration, Fig.1 gives 2 elements on 2 layers. Each circle is a RF. Bold circles form an element. The centers of all elements are denoted by plus signs. Note that in this figure as well as eleswhere in this paper, each unit represents one pixel.

The responses of all RFs within one element are collected to be a feature vector.
In our notation, the $j$th component of a feature vector can be written as ${\bf R}^j_{lm}$, which indicates the response of $l$th element on $m$th layer. 
$j$ is the RF index which ranges from $1$ to $19$.

\begin{figure}
\centering{
\includegraphics[width=.47\textwidth]{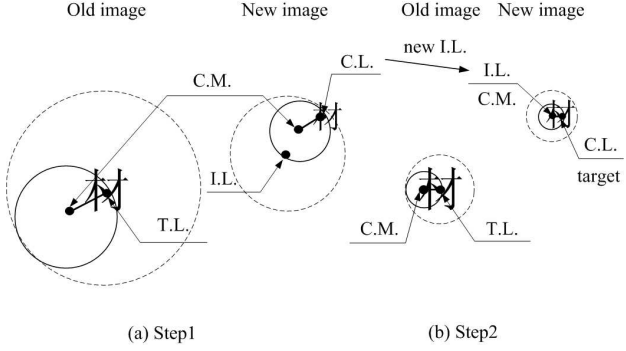}}
\caption{2 steps of matching. The solid circles represent matched elements while the dashed circles the whole layer. At each step, C.L. is calculated so that the relative position of C.M. and C.L. in the new image is the same as that of C.M. and T.L. in the old image up to a scale factor. C.L. is the initial of corrected location, C.M., center of matched element, T.L., training location, I.L., initial location, and F.L., final localized location.}
\label{fig_step}
\end{figure}

To eliminate local contrast change, we normalize the gabor intensity and map it to $16$ integers.
Precisely, the feature vector on $m$th layer is obtained as follows,
\begin{eqnarray}
{\bf R}^{j}_{lm}=ceil(16\times\frac{G^j_{im}-min(G_{im})}{max(G_{im})-min(G_{im})}),
\end{eqnarray}
where $ceil$ rounds a number up to the nearest integer, $G^j_{im}$ is the output of the convolution of Gabor filters $\phi_{{f_0}_m,\theta,\sigma_m}$ within jth RF, and the normalization (so the max and min operation) is taken over all RFs within the area $l$th layer covered. 
The same normalization and mapping procedure is followed on all layers so that we can compare the feature vectors across scales.

\section{Feature Matching and Point Localization}

In a reference image, some key points are first selected for training using some saliency based algorithms.
For recognition, we need to find those points in a new image that correspond to the selected points.

It is inefficient to search all possible locations and scales in the image.
Because of the great benefit to reduce non-relevant regions from pre-selection stage, we first use a saliency based algorithm to select some subregions and build our search model there.

The idea for searching is based on the multi-scaled representation of an object. Matching of features at different scales yields different resolutions. The matching of the larger layers works as coarse scanning and the matching of smaller layers works as fine localization. Suppose we want to localize a feature to the resolution of $r\times r$ in a $A\times B$ image. The computation complexity for the whole scanning is $\frac{ABS}{r^2}$, where $S$ is the number of scales at which the object is supposed to be recognizable. Using our system with $M$ layered representation, we can reduce the complexity to $\frac{AB\log_{\sqrt{2}}S}{r^2\sqrt{2}^M\sqrt{2}^M}$, with $\sqrt{2}$ being the ratio of the sizes of 2 adjacent layers. The computational complexity has been reduced about $2^M$.

The searching is done in 2 steps. 

First, the matching of the largest layer gives a coarse scanning. For a chosen $M$ layered mask (assuming that we want to detect an object on $M$ scales), we only get the feature vectors of the largest layer for a coarse scanning by putting the $M$th layer at an initial starting point. Then we match the feature vectors to the template in a nearest neighbor manner. The similarity of 2 feature vectors are defined to be, 
\begin{eqnarray}
Sim({\bf R}, {\bf R'})=exp(-\sum_{j}|{\bf R}^{j}_{lm}-{\bf R'}^{j}_{l'm'}|),
\end{eqnarray}
which takes value from $0$ to $1$. 
Once the nearest neighbors are found, for example, to be $l$th element on $M$th layer of the new image and $l'$th element on $m'$ of the template. Bearing the idea that the matched elements capture the same region of 2 images, we know that the scale of the new image is $r^{M-m'}$ relative to that of the original image from which we form our template. Also, we know that the matched elements should have the same relative position relative to the center of the mask weighted by the scale if the initial position is correctly placed. If the indices of the two matched elements are different $l\neq l'$, the center of the mask is mis-placed in the new image. We should move the mask to ${\bf X}_l-r^{M-m'}{\bf Y}_l'$, where ${\bf X}_l$ is the coordinate of the center of element $\{l,M\}$, and ${\bf Y})l'$ is the relative position of element $\{l',m'\}$ relative to the center of the mask in the template. 

Next, at the corrected location, feature vectors of the smallest layer (if $M$ is very large, we may need larger layer as well as the 
smallest one) are obtained to match the corresponding layer of the template, from which the target is localized.

\begin{figure}
\centerline
{\subfigure[Original Image]{\includegraphics[width=.28\textwidth]{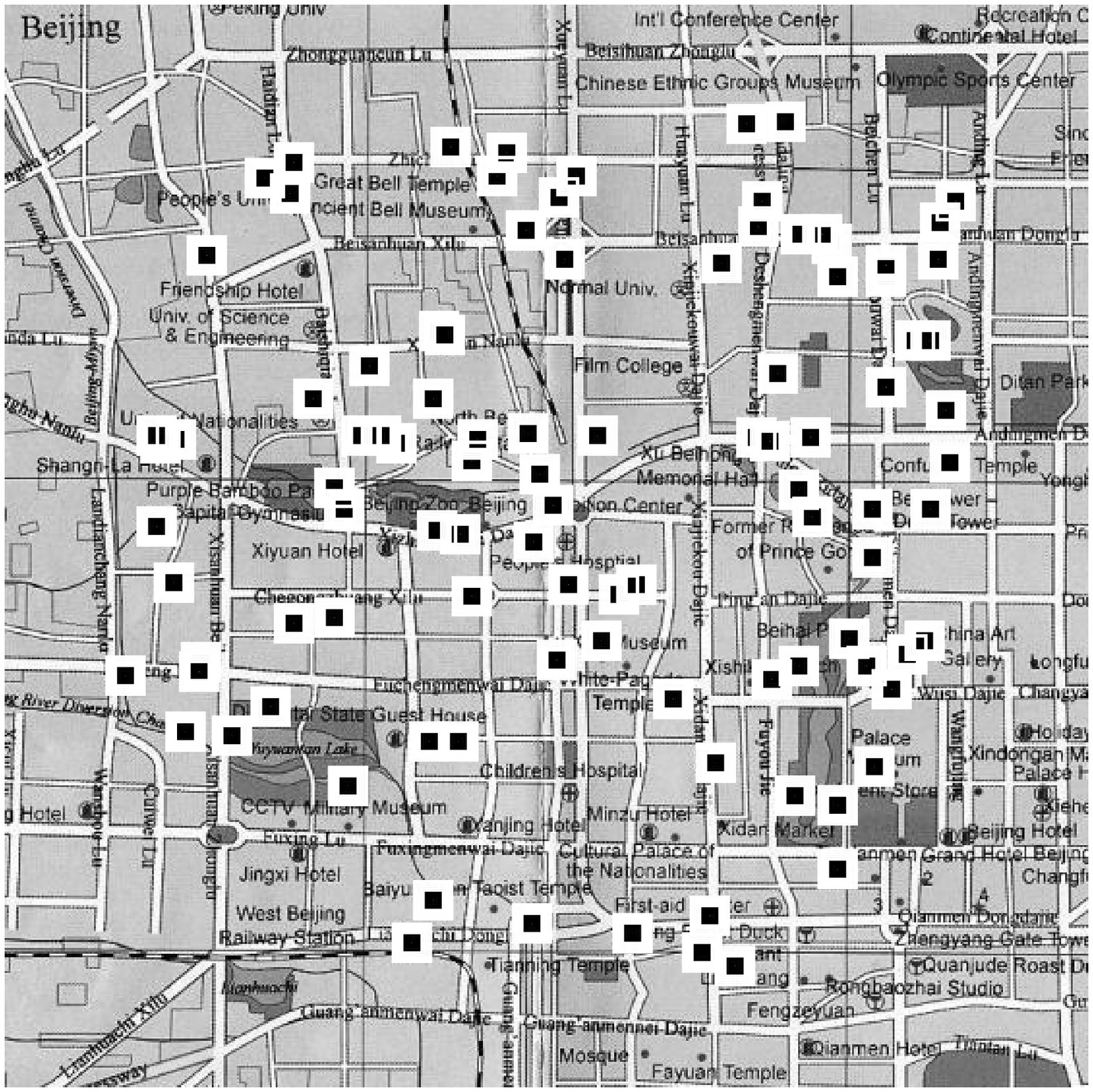}
\label{fig_map1}}
\hfil
\subfigure[Distorted Image]{\includegraphics[width=.23\textwidth]{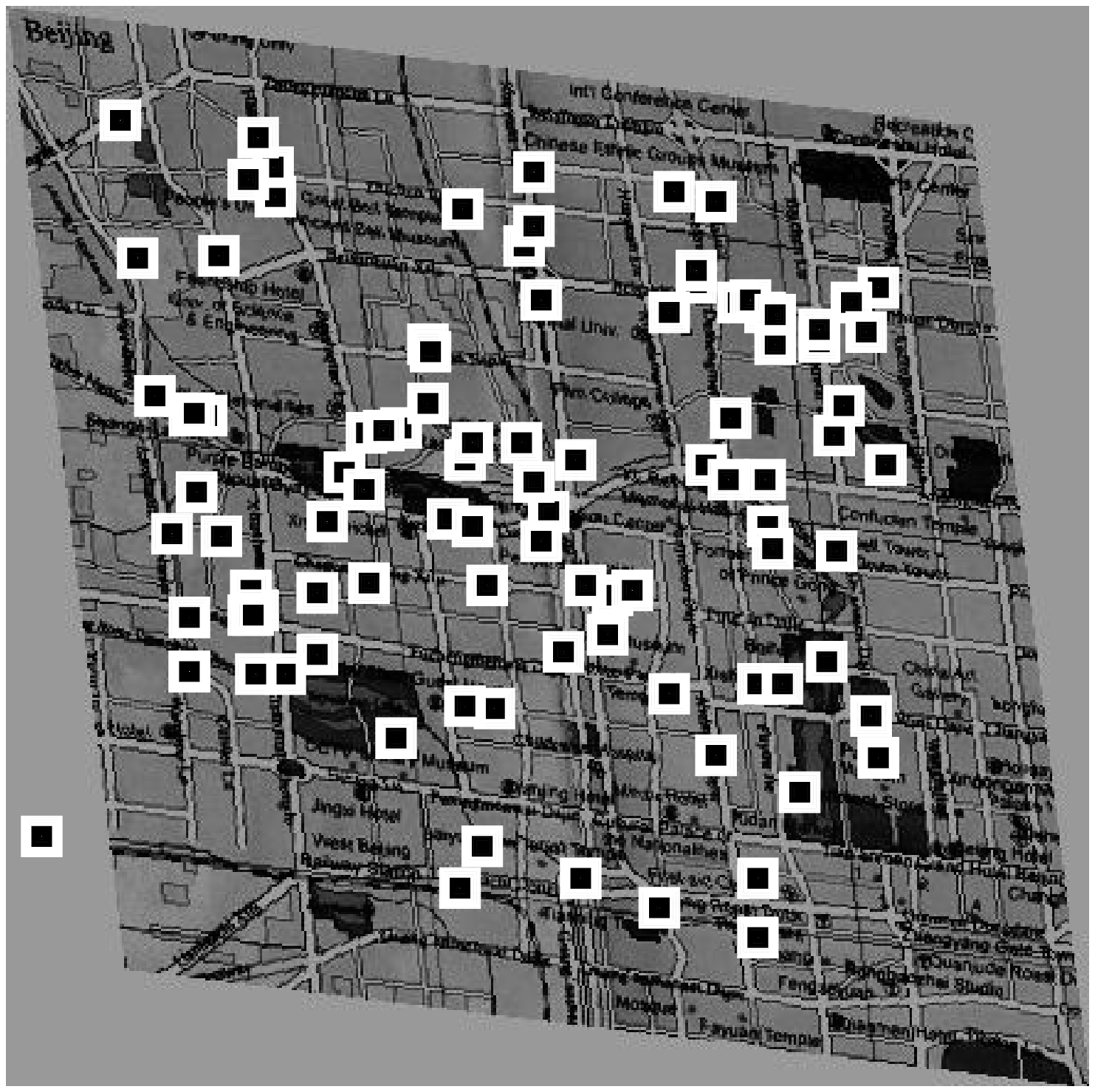}
\label{fig_map2}}}
\caption{An image and its transformed image by skewing, scaling, rotation, addition of pixel noise and change of brightness and contrast. }
\label{fig_map}
\end{figure}

The algorithm to match a feature in a new image with the template can be listed as,
\begin{itemize}
\item Pre-select some interesting subregions in a new image with saliency based algorithm.
\item Choose the central point from each subregion and obtain the feature vectors of the largest layer.
\item Find the nearest neighbor of the largest layer from the template.
\item Calculate the relative scale of two images and new initial location.
\item Localize the target from the new location with the smallest layer.
\end{itemize}

As an example, we demonstrate how this works using a Chinese word. First, the template is formed in the image centered around a Chinese word which is located at the training location (T.L.) $[197,259]$.

The localization is finished in 2 steps as shown in Fig.2. The first step, we put the 5th layer near the target at initial location (I.L.) 
in the new image (so that our system can recognize the same Chinese word from 2 times large to 2 times smaller than the original training word). We collect responses from all the receptive fields on the 5th layer. Two matched elements (C.M.)  are found to be ${l=12, m=5}$ located at $[137, 293]$ 
in the new image and ${l'=15, m'=6}$ at $[120, 203]$ in the old image. 
The scale of the new image is identified to be $\sqrt{2}^{(m-m')}$ times larger than the old image. 
Since $l \neq l'$, the corrected location (C.L.) is calculated to be $[120, 203]-([137, 293]-[197, 259])/\sqrt{2}=[162, 179]$. The second step simply uses the 1st layer at the corrrected location calculated from step 1 and repeat the above procedure. The final localized point is $[182, 179]$ which exactly corresponds to the trained point with high accuracy.

Searching in each subregion yields one candidate for the object's location in the new image. In a new image that has multiple objects, a threshold is useful to identify the objects. If the prior information is known that there is only one match in a new image, we can simply choose the best one which proves to be effective in the map mapping experiment (the false alarm rate for our representation is small).
The final evaulation of these candidate locations is made using the similarity function comparing one element or a set of elements (cf. to the experimental section for the difference of results) centering around each candidate location. 
The evaluation function for each located candidate is defined to be, 
\begin{eqnarray}
Eval=\prod_i^T{Sim({\bf R}_i, {\bf R}'_i)},
\label{evalu}
\end{eqnarray}
where $i$ goes from 1 to the number of elements $T$ we want to use, and ${\bf R_i}$ and ${\bf R'_i}$ are the corresponding feature vector for that element, respectively, in the new image and in the old image.

\section{Experimental examples}

\begin{table}

\label{table1}
\centering{
\begin{tabular}{|l|c|r|}
\hline
Image Transformation & rate1 \% & rate2 \% \\
\hline
A.Increase contrast by 1.2 & 97 & 96 \\
B.Decrease intensity by 0.2 & 95 & 92 \\
C.Rotate by 10 degress & 91 & 81 \\
D.Scale by 0.7 & 99 & 99 \\
E.Scale by 0.5 & 99 & 97 \\
F.Add 10\% pixel noise & 98 & 98 \\
G.Skewed by 7 degrees & 95 & 88 \\
H.Scale by 1.5 & 95 & 92 \\
I.All of A,B,D,F,G & 88 & 83 \\
\hline
\end{tabular}
\caption{For various image transformations applied to the original image, the table gives the percentage of matching within 8 pixels away from the true target calculated according to the transformations. }}
\end{table}
Fig. 3 shows an example of points matching. The first image is a reference image and the second image is a transformed image.
We random choose 100 points in the first image and search those points on the transformed image. We divide the image into 150 by 150 subregions for a 800 by 800 picture. In each region, we localize a candidate and the best of these candidates is taken to be the recovered point. If the recovered point is less than 8 pixels away from the actual location calculated from the transformation, a matching is found. Table 1 shows the results. The first column of data gives the matching rate when all elements ($T=19$ in Equa. 4) on the smallest layer are used for the evaluation while the last column gives the matching rate when the evaluation is done only on the matched element. Our algorithm is weak in the detection of rotation related transforms. We did nothing in the representation to make it rotationally invariant, although we can achieve this simply by shifting the Gabor intensity matrices \cite{gabor1}. This is justified because the human visual system is not rotationally invariant.

The second experiment is designed for recognition. In the trained image, 5 key points are selected on an object for training, which are then recovered in the recognition image.

\begin{figure}
\centerline
{\subfigure[Trained Image]{\includegraphics[width=.25\textwidth]{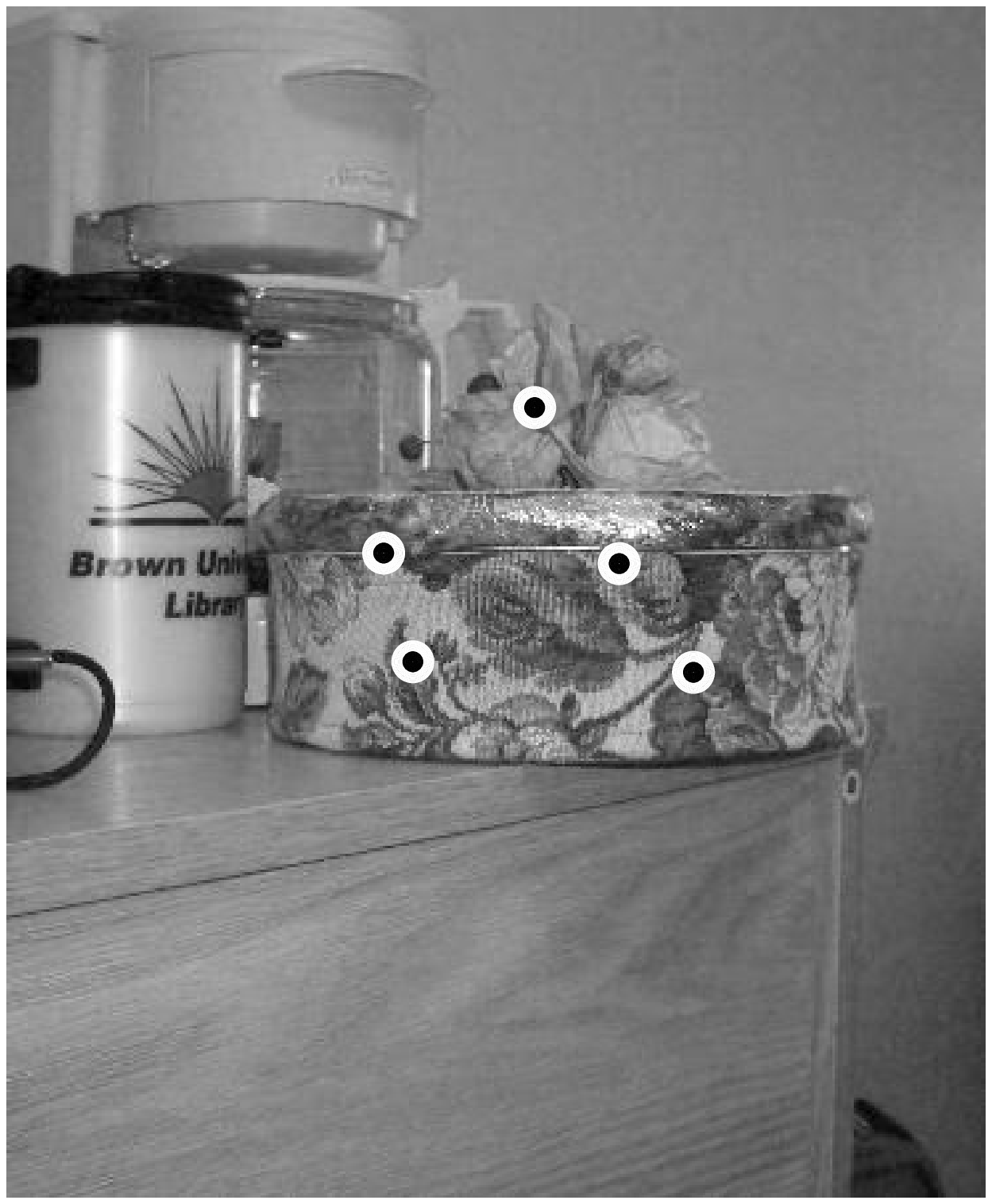}
\label{fig_rabbit1}}
\hfil
\subfigure[Recognition Image]{\includegraphics[width=.25\textwidth]{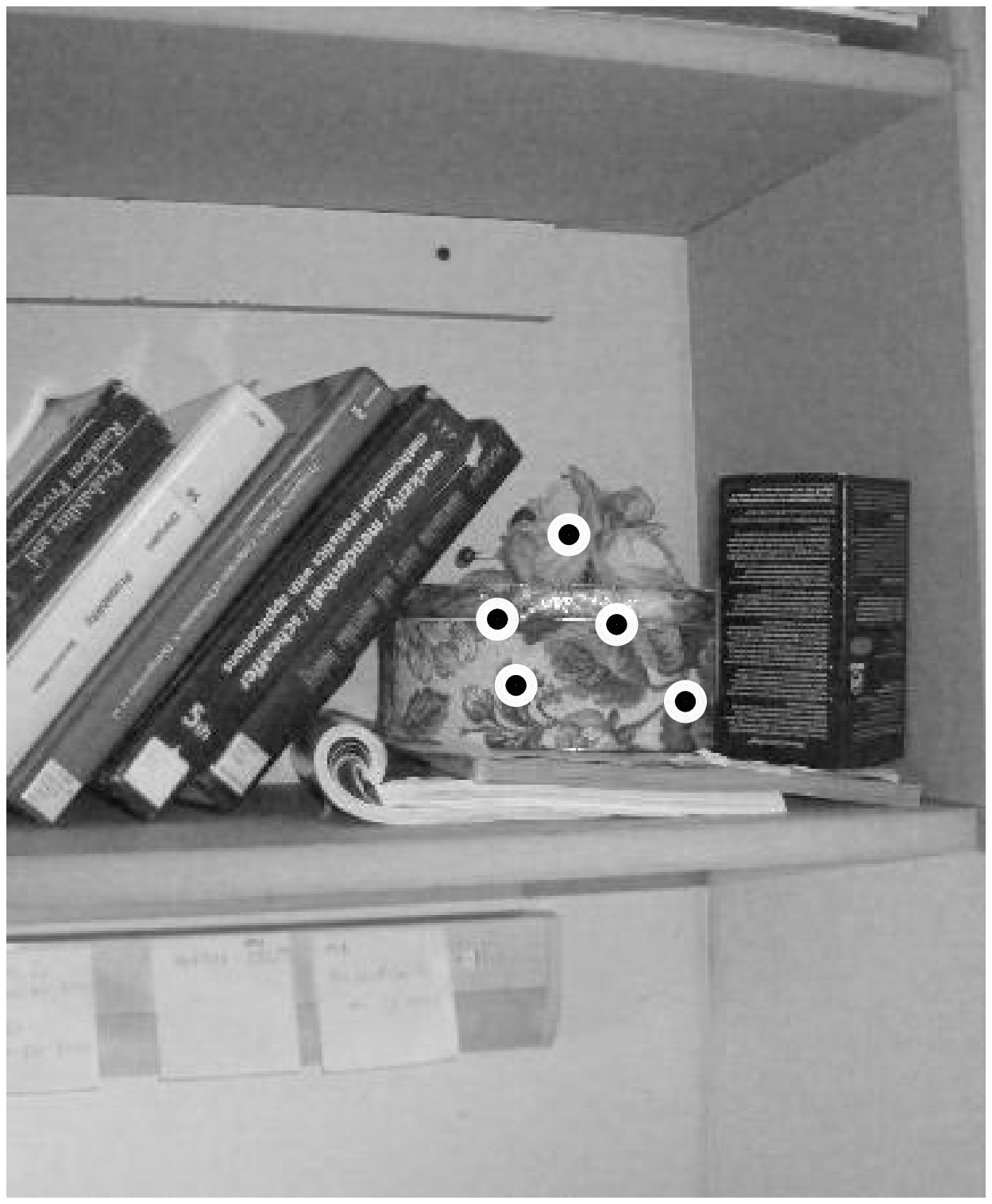}
\label{fig_rabbit2}}}
\caption{(a) 5 points selected on an object in the trained image and (b) 5 recovered points in the recognition image}
\label{fig_rabbit}
\end{figure}

\section{Conclusion and Discussion}
In this paper, we present a feature extractor and an associated searching model. The motivation to design our model is based on the observation of human perception. 
We look at an image with attention to a specific object while ignoring others and we use a strategy to search from coarse to fine. To build such a model, a hierarchical representation of an object is given at different scales. Coarse scanning is done through the matching of the larger scale and precise localization is conducted through the matching of the smaller scale.
Experimental results justify the proposed model in its effectiveness and efficiency to localize features with very high accuracy. 

In this short paper, we do not explore too much the bottom-up algorithms for key points selection at learning stage. 
However, our proposed model can be potentially combined with those techniques to improve the performance of the whole system.


\bibliographystyle{latex8}
\bibliography{icpr}

\end{document}